\documentclass[journal]{IEEEtran}
\usepackage{amsmath,epsfig}
\usepackage[noadjust]{cite}
\usepackage{cite}
\usepackage{amssymb}
\usepackage{amsfonts}
\usepackage{mathtools}
\usepackage{bbm}
\usepackage{pifont}

\usepackage{upgreek}
\usepackage{float}
\usepackage{graphicx}
\usepackage{subcaption}  
\usepackage{multicol}
\usepackage{multirow}
\usepackage{verbatim}
\usepackage{soul}
\usepackage{booktabs}

\usepackage[hidelinks]{hyperref}
\usepackage{amsmath,amssymb,amsfonts}
\usepackage[linesnumbered,lined,boxed,commentsnumbered,ruled]{algorithm2e}
\usepackage[usenames,dvipsnames]{xcolor} 
\usepackage{orcidlink}

\usepackage{adjustbox}
\usepackage{makecell}
\usepackage{comment}

\usepackage[dvipsnames]{xcolor}
\usepackage{soul}
\sethlcolor{yellow}

\DeclareMathAlphabet\mathbfcal{OMS}{cmsy}{b}{n}

\begin{document}

\title{FedX: Explanation-Guided Pruning for Communication-Efficient Federated Learning in Remote Sensing  }
%
\author{ Bar{ı}\c{s} B\"{u}y\"{u}kta\c{s}$^*$ \orcidlink{0000-0002-8240-8784}, 
         Jonas Klotz$^*$ \orcidlink{0009-0003-3297-2365}~\IEEEmembership{Member,~IEEE},
         Beg\"{u}m Demir \orcidlink{0000-0003-2175-7072}~\IEEEmembership{Senior Member,~IEEE}%

        \thanks{Barı\c{s} B\"{u}y\"{u}kta\c{s}, Jonas Klotz, and Beg{\"u}m Demir are with the Faculty of Electrical Engineering and Computer Science, Technische Universit\"at Berlin, 10623 Berlin, Germany, also with the BIFOLD - Berlin Institute for the Foundations of Learning and Data, 10623 Berlin, Germany.
        Email: \mbox{baris.bueyuektas@tu-berlin.de}, \mbox{j.klotz@tu-berlin.de}, \mbox{demir@tu-berlin.de}.
        }
        
}

\maketitle
\def\thefootnote{*}\footnotetext{These authors contributed equally to this work}

\begin{abstract}

Federated learning (FL) enables the collaborative training of deep neural networks across decentralized data archives (i.e., clients), where each client stores data locally and only shares model updates with a central server. This makes FL a suitable learning paradigm for remote sensing (RS) image classification tasks, where data centralization may be restricted due to legal and privacy constraints. However, a key challenge in applying FL to RS tasks is the communication overhead caused by the frequent exchange of large model updates between clients and the central server. To address this issue, in this paper we propose a novel strategy (denoted as FedX) that uses explanation-guided pruning to reduce communication overhead by minimizing the size of the transmitted models without compromising performance. FedX leverages backpropagation-based explanation methods to estimate the task-specific importance of model components and prunes the least relevant ones at the central server. The resulting sparse global model is then sent to clients, substantially reducing communication overhead. We evaluate FedX on multi-label scene classification using the BigEarthNet-S2 dataset and single-label scene classification using the EuroSAT dataset. Experimental results show the success of FedX in significantly reducing the number of shared model parameters while enhancing the generalization capability of the global model, compared to both unpruned model and state-of-the-art pruning methods. The code of FedX is publicly available at https://git.tu-berlin.de/rsim/FedX.

\end{abstract}


\section{Introduction}
\label{intro}

Remote sensing (RS) image classification plays a crucial role in a wide range of RS applications \cite{li2024deep,la2023learning,zhang2023single}. In recent years, deep learning (DL) techniques have significantly advanced the performance of RS image classification by enabling the extraction of discriminative features to represent the complex content of images \cite{sumbul2020deep,fang2025deformable}. 
Most existing DL techniques typically require that all training data to be collected and stored on a single central server for model training. However, RS images are often distributed across decentralized image archives (i.e., clients) \cite{buyuktas2024federated}.
Due to bandwidth and storage limitations, consolidating all RS data on a central server is challenging \cite{chen2021communication}. This challenge is further compounded due to the continuous increase in the volume and resolution of RS data. In addition to data storage constraints, there are also growing concerns about data sensitivity and privacy in decentralized image archives. In RS, especially with the continued advancement of high-resolution imagery, there is an increasing risk of capturing sensitive information \cite{mcamis2024over}. As an example, satellite images used in urban monitoring can reveal detailed views of private properties, such as building layouts and vehicles, thereby raising ethical and legal concerns related to personal privacy. Additionally, while RS images alone may not raise privacy concerns, their combination with other data sources can lead to situations that pose significant privacy risks. As an example, the fusion of RS images and socio-economic data (e.g., income levels, education, health statistics etc.) can significantly increase the risk of re-identification. This enables highly detailed profiling of specific communities or even individuals. In addition, legal regulations often prohibit the sharing of RS images that involve the monitoring of sensitive sites such as borders or critical infrastructure \cite{avtar2021remote}. These challenges highlight the need for a learning approach that enables collaborative model training across distributed RS image archives, where RS images remain unshared due to privacy concerns and legal constraints.

Federated learning (FL) addresses this need by introducing a decentralized training strategy in which clients collaboratively train models by sharing only model updates with a central server. This enables the joint model training without compromising data confidentiality. Moreover, by keeping data localized, FL alleviates the need for centralized data storage and ensures compliance with legal data handling requirements. However, one notable limitation of FL is the communication overhead associated with frequent exchanges of model updates between clients and the central server. In each training round, model parameters are required to be transmitted back and forth, which can be challenging in environments with limited bandwidth or a large number of participating clients. This is particularly relevant in RS, where clients may include edge devices such as unmanned aerial vehicles or satellites with constrained communication capabilities. This overhead can slow down training and introduce latency, making it challenging to deploy FL in real-time or resource-constrained environments. While the communication overhead in FL has been extensively studied in the computer vision (CV) community (see Section \ref{lit} for a detailed review), relatively few works have addressed this limitation in the context of RS \cite{duong2024leveraging,yang2024communication,song2023efficient, li2024feddiff,henna2026spatiotemporal,zhang2023federated}. 

In \cite{duong2024leveraging}, a feature-based FL framework is proposed for RS image classification, where clients transmit low-dimensional embeddings or pseudo-weights instead of full models. To this end, image representations are extracted from RS images and shared with the server to enable collaborative training with reduced communication. In \cite{yang2024communication}, a satellite-ground FL framework is introduced to address bandwidth limitations in spaceborne RS. Accordingly, model updates are compressed through progressive weight quantization before being transmitted from satellites to ground stations. In \cite{song2023efficient}, a pruning-based FL method is proposed for RS using unmanned aerial vehicles, where clients apply one-shot filter pruning to remove less important convolutional filters prior to communication. In \cite{li2024feddiff}, a diffusion model–driven FL framework (denoted as FedDiff) is proposed for secure multi-modal RS data fusion across multiple clients. To this end, a dual-branch diffusion architecture with a lightweight communication module is embedded into the FL framework to enable communication-efficient and privacy-preserving multi-modal knowledge exchange. In \cite{henna2026spatiotemporal}, a federated generative adversarial framework for privacy-preserving synthesis of multi-source RS time series (denoted as FedScoreGAN) is proposed under non-IID data. The framework integrates score-based generative modeling with federated aggregation and incorporates a communication-aware design via parameter sparsification. In \cite{zhang2023federated}, a federated DL scheme with prototype matching (denoted as FedPM) is proposed to enable training across multiple clients with heterogeneous data. The scheme enforces class-wise prototype alignment during local optimization to reduce feature distribution discrepancies across clients. Prototype deviations are then leveraged to guide attention-weighted global aggregation, while sparse ternary compression is integrated to alleviate communication overhead. To mitigate communication overhead in FL, existing methods in CV are commonly classified into three main categories: 1) model compression \cite{reisizadeh2020fedpaq,li2024fedsparse,zhao2023aquila,tsouvalas2024communication}; 2) knowledge distillation \cite{lin2020ensemble,han2024fedal,zhang2024fedgmkd,wu2023fedict}; and 3) pruning \cite{jiang2022model,zhu2023fedlp,huang2023distributed,zhou2024personalized}. Model compression and knowledge distillation are widely used strategies for reducing communication overhead in FL. Model compression-based methods address communication overhead by quantizing the size of model updates before transmission. However, these techniques may lead to a drop in model performance due to the removal or approximation of important parameters \cite{dantas2024comprehensive}. They can also introduce computational overhead, as clients and servers often need to perform additional encoding and decoding steps during training. Knowledge distillation-based methods reduce communication by transmitting soft predictions instead of full model parameters, allowing the server to aggregate knowledge from clients without sharing full models. However, these approaches rely on student models that may fail to match the performance of the teacher models, especially for complex tasks. Moreover, knowledge distillation may still leak sensitive information through the shared logits \cite{xiao2024privacy}. Pruning strategies offer advantages over both knowledge distillation and model compression by directly reducing model size without requiring separate student models or extra encoding steps. As a result, pruning maintains model structure and minimizes additional computational overhead.


In this paper, we focus on pruning-based methods and investigate the effectiveness of explanation-guided pruning in the context of FL for RS. 
To address this, we propose a strategy that utilizes backpropagation-based explanation methods to effectively reduce the communication overhead in FL for RS applications. We specifically focus on explanation-guided pruning in FL because magnitude-based and locally driven pruning methods can be less reliable for identifying globally relevant components. After aggregation across heterogeneous clients, updates from different data distributions may partially cancel each other out, causing parameters that are important for some clients to have small magnitudes in the global model despite carrying task relevant information. To overcome this limitation, our strategy leverages the explanation signals computed at the central server to assess the relevance of model parameters. Based on the relevance scores, the server performs pruning by removing less important neurons or filters from the global model. The pruned global model is then sent back to the clients for the next round of local training. By performing pruning methods at the server side, our strategy reduces the size of the global model transmitted in each round while preserving task-relevant information. This results in significantly lower communication costs, making the strategy particularly well-suited for large-scale, bandwidth-limited, and privacy-sensitive RS environments. It is worth noting that a preliminary version of our strategy was briefly introduced in \cite{klotz2025communication} with limited experimental evaluation. In this paper, we significantly extend that work by presenting a comprehensive description of the proposed strategy and conducting extensive experiments in the context of multi-label scene classification using the BigEarthNet-S2 dataset and single-label scene classification using the EuroSAT dataset. In addition, we extend the proposed methodology by replacing global pruning with a layer-wise pruning configuration. This change addresses the limitation of global pruning, which disproportionately removes components from deeper layers due to lower average relevance scores caused by relevance conservation. Layer-wise pruning configuration addresses this issue by applying thresholds within each layer, resulting in more balanced pruning and improved model robustness under high sparsity. To the best of our knowledge, it is the first FL strategy that utilizes explanation-guided pruning to reduce communication overhead while preserving model performance in image classification tasks.

The rest of this article is organized as follows. Section II presents the related works. Section III introduces the proposed strategy. Section IV describes the considered RS image archives and the experimental setup, while Section V provides the experimental results. Section VI concludes our article.

\section{Literature Review}
\label{lit}

As mentioned in Section \ref{intro}, there are only few studies that have investigated communication-efficient FL in the context of RS. However, this topic has been widely studied in CV and existing studies can be grouped into three different categories: 1) model compression-based algorithms; 2) knowledge distillation-based algorithms; and 3) pruning-based algorithms. In the following subsections, we review the communication-efficient FL algorithms.

\subsection{Model Compression-Based FL Algorithms}

Model compression-based FL algorithms aim to reduce communication overhead by compressing model updates through techniques such as quantization and sparsification. Compression can be applied on the client side to reduce the size of the local model or on the server side to minimize the size of the global model sent back to clients. In \cite{reisizadeh2020fedpaq}, a communication-efficient FL algorithm (denoted as FedPAQ) is proposed to combine periodic model averaging with quantized communication. To this end, clients perform multiple local training steps before sending their updates, and the updates are quantized using low-bit encoding schemes to further reduce transmission size. In \cite{li2024fedsparse}, a sparse communication framework (denoted as FedSparse) is proposed to enhance communication efficiency in FL through sparsity-aware local optimization. To achieve this, FedSparse incorporates two components: 1) resource optimization proximal term that penalizes excessive updates to reduce transmission size; and 2) importance-based regularization weighting mechanism that assigns higher sparsity to less important parameters. This formulation enables clients to jointly optimize empirical risk and communication cost. In \cite{zhao2023aquila}, an adaptive quantization framework (denoted as AQUILA) is proposed to integrate quantization into the device selection strategy for more effective participation of clients. To this end, AQUILA introduces a device selection method that prioritizes updates based on their quality and relevance to the global model, using the exact global model stored on each device. It also presents an adaptive quantization criterion that removes the need for manual tuning while ensuring convergence. In \cite{tsouvalas2024communication}, a model compression framework (denoted as FedCompress) is proposed to reduce uplink and downlink communication by combining dynamic weight clustering. Accordingly, clients compress local models by grouping similar weights during training, and transmit these clustered representations instead of full model parameters. Despite their benefits, model compression-based FL algorithms may remove important parameters during quantization or sparsification. They also introduce additional computational overhead due to the need for encoding and decoding at both client and server sides. These limitations may limit the overall effectiveness of model compression, motivating the exploration of alternative strategies.

\subsection{Knowledge Distillation-Based FL Algorithms}

Knowledge distillation-based FL algorithms aim to reduce communication overhead by exchanging soft predictions or intermediate representations instead of full model parameters. In this setting, clients can act as teacher models by sharing their output logits with the central server, which then distills the knowledge into a student model. This allows the server to learn from client knowledge without requiring direct access to model weights or local data. In \cite{lin2020ensemble}, a knowledge distillation framework (denoted as FedDF) is proposed to aggregate knowledge from diverse local models without sharing the parameters. To achieve this, clients send soft predictions on a public dataset to the server, which then uses these predictions to distill a global model. In \cite{han2024fedal}, a federated knowledge distillation framework (denoted as FedAL) is introduced to align client predictions through adversarial learning. To this end, the server functions as a discriminator that guides local training by enforcing output consistency across clients through an adversarial learning process. This min-max game reduces divergence in model predictions caused by heterogeneous data distributions. Additionally, a less-forgetting regularization is applied during both local training and global knowledge transfer to mitigate the effects of catastrophic forgetting, ensuring stable and effective knowledge sharing among clients. In \cite{zhang2024fedgmkd}, a prototype-based FL framework (denoted as FedGMKD) is introduced to enhance personalization and generalization through knowledge distillation and discrepancy-aware aggregation. To this end, clients generate prototype features using Gaussian Mixture Models and transmit soft predictions instead of raw data. The server aggregates these prototypes by weighting client contributions according to data quality and quantity, enabling effective knowledge transfer without relying on public datasets or generative models. In \cite{wu2023fedict}, a federated multi-task knowledge distillation framework (denoted as FedICT) is proposed to enable bi-directional knowledge transfer between clients and the server without requiring a public dataset. To achieve this, FedICT introduces two key components: 1) federated prior knowledge distillation that uses local data distribution information to improve the fitting of client models; and 2) local knowledge adjustment that refines the server’s distillation loss to better align transferred local knowledge with generalized global representations. Knowledge distillation-based FL algorithms help reduce communication by sharing soft predictions instead of full models. However, student models may not perform well in complex tasks, and shared logits can leak private information. These challenges motivate the use of pruning as an alternative method to address communication overhead in FL.

\subsection{Pruning-Based FL Algorithms}

Pruning-based FL algorithms aim to reduce model complexity by identifying and removing less important parameters such as individual weights, neurons, or convolutional filters. In \cite{li2021lotteryfl}, a pruning-based FL framework (denoted as LotteryFL) utilizes the Lottery Ticket Hypothesis to identify sparse subnetworks on each client. Each client independently learns a lottery ticket network, and only these client-specific subnetworks are communicated between the server and the participating clients during training. In \cite{jiang2022model}, a pruning-based FL framework (denoted as PruneFL) is introduced to reduce communication and computation costs by adaptively pruning local models during training. To this end, PruneFL estimates the importance of model weights based on their magnitude and prunes less significant parameters on the clients before transmitting updates to the server. In \cite{bibikar2022federated}, a federated dynamic sparse training framework (denoted as FedDST) maintains a fixed global sparsity throughout training. Each client applies layer-wise magnitude-based pruning at every communication round and restores pruned weights using random initialization or gradient-based criteria. In \cite{zhu2023fedlp}, a layer-wise pruning algorithm (denoted as FedLP) is proposed to address the communication overhead in FL by systematically pruning entire layers or layer components. Accordingly, FedLP employs a unified optimization framework that jointly considers pruning ratio and layer selection, allowing each client to independently adapt the pruned model based on local resource constraints. In \cite{huang2023distributed}, a distributed pruning framework (denoted as FedTiny) is introduced for FL that generates highly compressed models suitable for memory- and computation-constrained devices. To this end, FedTiny integrates two core components: an adaptive batch normalization selection module to reduce pruning bias caused by heterogeneous local data distributions, and a lightweight progressive pruning module that incrementally determines layer-wise pruning policies under strict resource constraints. In \cite{long2023feddip}, a dynamic pruning framework (denoted as FedDIP) combines extreme pruning with incremental regularization. During local training, each client performs dynamic pruning by removing low-magnitude weights and then progressively applies an $L_2$ regularization to stabilize the pruned structure. In \cite{zhou2024personalized}, a personalized FL algorithm with adaptive pruning is proposed by introducing a two-stage approach in which the initialization phase rapidly identifies sparse subnetworks, and the adaptive pruning phase customizes models to local data. This design enables efficient training while reducing both communication and computational overhead on edge devices. Existing pruning methods rely on simple heuristics such as weight magnitude and offer limited insight into which parameters are removed. This can result in the loss of task-relevant information. To address this, we propose an explanation-guided pruning strategy that uses relevance scores to retain the most informative components during pruning.

\begin{figure*}[t]
\centering
\centerline{\includegraphics[width=2\columnwidth,keepaspectratio]{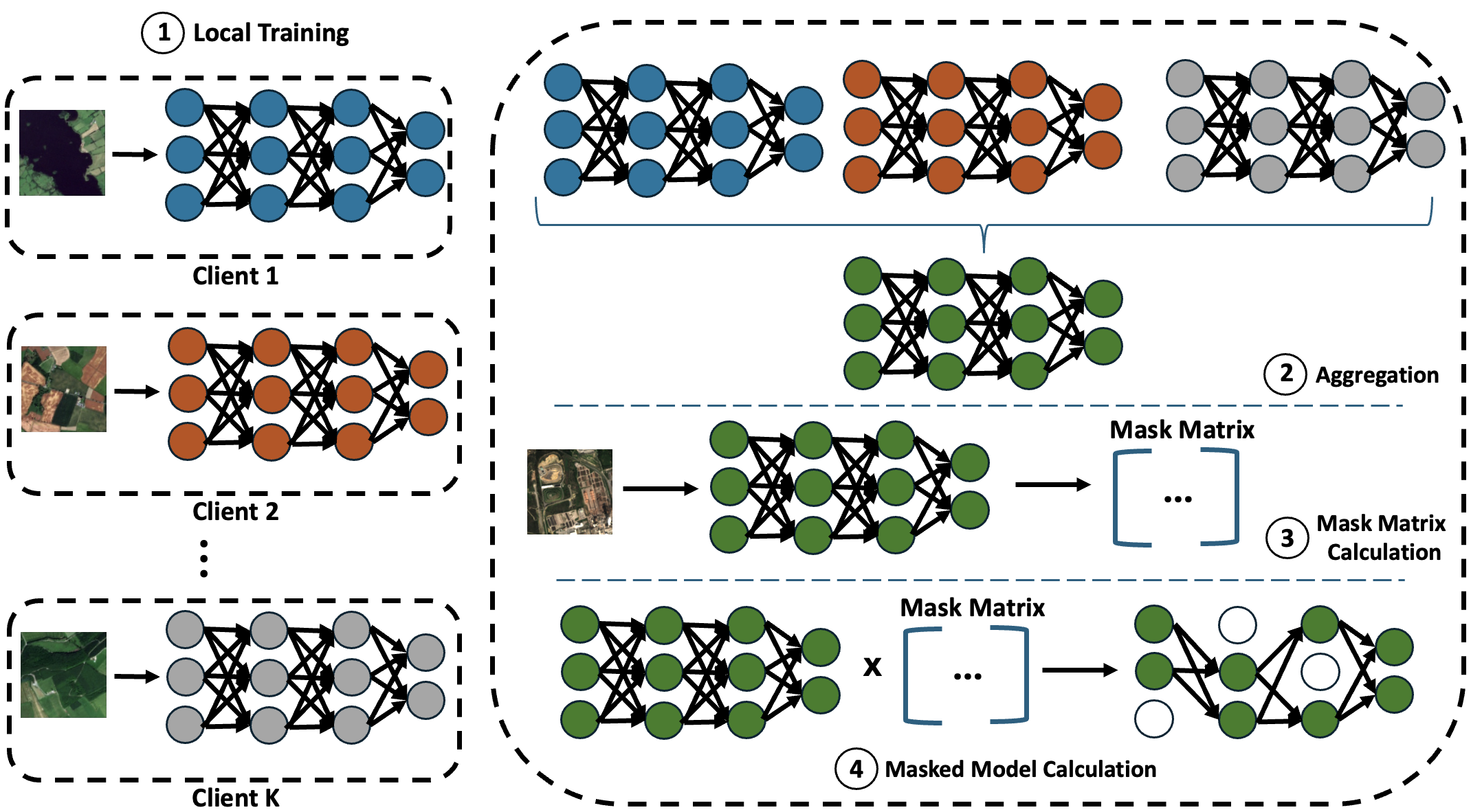}}
\caption{Overview of FedX. (1) Clients perform local training on their private RS data. (2) The central server aggregates the locally trained models. (3) Reference RS data is used to compute explanation-based relevance scores, from which a global mask matrix is derived. (4) The global model is pruned using the mask matrix, resulting in a sparsified model distributed to clients for the next training round.  }
\label{figure:strategy}
\end{figure*}

\section{Proposed Explanation-Guided Pruning Strategy for FL}

To mitigate the communication overhead in FL for RS image classification, we propose FedX, a pruning strategy that integrates explanation-based techniques to reduce the size of the global model updates exchanged between clients and the central server. The proposed strategy ensures that only the most relevant and informative parts of the global model are retained and distributed, thereby significantly decreasing communication costs without compromising model effectiveness. In the following subsections, we first formalize the FL problem and then present the explanation-guided pruning strategy. Fig. \ref{figure:strategy} shows an overview of the proposed strategy.

\subsection{Problem Formulation}

In the FL setting, we consider a set of $K$ clients, denoted as ${C_1, C_2, ..., C_K}$. Each client $C_i$ holds local training data $D_i$ consisting of $M_i$ labeled samples:

\begin{equation}
D_i = \left\{(\boldsymbol{x}_{i}^z, \boldsymbol{y}_{i}^z)\right\}_{z=1}^{M_i},
\end{equation}
where $\boldsymbol{x}_{i}^z$ represents the $z$-th training sample and $\boldsymbol{y}_{i}^z$ is its associated label. In the context of image classification, $\boldsymbol{y}_{i}^z$ may be a single label or a multi-label annotation depending on the classification task. We assume that the local training data is private and non-shared among clients, in accordance with the privacy-preserving principles of FL. Each client $C_i$ maintains and trains a local model $\phi_i$, initialized with global model parameters. The objective of client-side training is to minimize the empirical risk on the local data:

\begin{equation}
\label{eq:local_obj}
\begin{aligned}
\mathcal{O}_i (D_{i}; w_i)  
      &= \!\!\!\!\! \sum_{(\boldsymbol{x}_{i}^z, \boldsymbol{y}_{i}^z) \in D_i} \!\!\!\!\!
      \mathcal{L}(\phi_i(\boldsymbol{x}_{i}^z; w_i), \boldsymbol{y}_{i}^z),\\
     w_i^* &= \arg \min_{w_i} \mathcal{O}_i(D_i; w_i),
\end{aligned}
\end{equation}
where $\mathcal{L}$ denotes the task-specific loss function, which can be categorical cross-entropy for single-label classification or binary cross-entropy for multi-label classification. The optimal local parameters $w_i^*$ are obtained by minimizing $\mathcal{O}_i$ over the training set of the clients through local training. After local updates, the parameters $w_i^*$ are transmitted to the central server, where they are aggregated to update the global model parameters $w$ as follows:

\begin{equation}
\label{eq:agg}
w = \sum_{i=1}^{K} \alpha_i w_i^*,
\end{equation}
where $\alpha_i$ denotes the aggregation weight for client $C_i$, often proportional to the local data size. This iterative communication of model updates between clients and the server leads to considerable communication overhead, particularly in scenarios where model architectures are large and network bandwidth is limited. Our objective is to alleviate this communication overhead while preserving generalization capability of the global model by introducing a pruning mechanism at the server side. To this end, we investigate explanation-guided pruning, which leverages model interpretability to identify and remove parameters that contribute least to the predictions of the model. Backpropagation-based explanation methods (e.g., layer-wise relevance propagation (LRP) \cite{bach2015pixel} and integrated gradients (IG) \cite{sundararajan2017axiomatic}) are particularly well-suited for this purpose, as they are designed to trace prediction contributions through the network layers. In the proposed strategy, relevance scores guide the pruning process at the server, allowing the identification and removal of globally less important components. This results in a sparse yet informative global model that is subsequently distributed to all clients. The details of this pruning strategy are described in the following subsection.

\subsection{Explanation-Guided Pruning at the Server}

The main idea of the proposed pruning strategy is to systematically identify and remove globally redundant or less important components of the model based on their quantified task relevance. The global model $\phi^*$ is structurally decomposed into a set of $p$ discrete components $\{\psi_1, \psi_2, ..., \psi_p\}$. Each component $\psi_c$ corresponds to a structured unit of the network (e.g., a convolutional filter), enabling structured pruning and communication-efficient model updates. The granularity of these components is flexible and can be adapted to suit the model architecture and pruning objectives. Specifically, components may correspond to convolutional filters, neurons, or attention heads, each representing a meaningful functional block within the model. This modular representation allows for pruning at various levels of abstraction, from coarse-grained units (e.g., entire layers) to fine-grained elements (e.g., individual filters). To assess the relevance of each component, the server utilizes a reference set denoted as $D_\text{ref} = \{(\boldsymbol{x}^z, \boldsymbol{y}^z)\}_{z=1}^{M_{ref}}$, where $M_{ref}$ is the number of samples in the reference set. This dataset provides the necessary context for evaluating the contribution of individual components. To ensure that the relevance assessment remains both privacy-preserving and generalizable, $D_\text{ref}$ can be constructed using publicly available or proxy data sources. For each input sample $I_z \in D_\text{ref}$, the explanation-guided pruning methods are applied to propagate the prediction of the model backward through the network to compute relevance scores at the component level. Specifically, for component $k$ in layer $l$, the relevance score $R_k^{(l)}(I_z)$ is determined by the proportional redistribution of relevance from the subsequent layer:

\begin{equation}
R_k^{(l)}(I_z) = \sum_{j} \frac{v_{k,j}}{\sum_k v_{k,j}} R_j^{(l+1)}(I_z),
\end{equation}
where $v_{k,j}$ reflects the contribution of component $k$ to component $j$ in the next layer, and $R_j^{(l+1)}(I_z)$ is the relevance score at the higher layer. The relevance is computed in a layer-wise backpropagation manner from the output to the input layer using specific propagation rules that ensure relevance conservation throughout the network \cite{montavon_layer-wise_2019}. For the final layer $L$, the relevance $R_k^{(L)}(I_z)$ is initialized as the output score corresponding to the target class $c$. Following component-level relevance estimation, the next step aggregates relevance scores at the component level. This process is repeated for all samples in $D_\text{ref}$. Then, the average relevance score of component $\psi_c$ across $D_\text{ref}$ is computed to obtain its global relevance estimate as follows:

\begin{equation}
\bar{R}{\psi_c} = \frac{1}{M_{ref}} \sum_{z=1}^{M_{ref}} R_{\psi_c}(I_z).
\end{equation}
Let $\mathbb{L}$ denote the set of all layers in the model. For each layer $l \in \mathbb{L}$, let $\Psi^{(l)}$ be the set of components in that layer. We apply pruning independently in each layer by defining a binary pruning mask $\mathcal{M}(\psi_c)$ for every component $\psi_c \in \Psi^{(l)}$ as follows:
\begin{equation}
\mathcal{M}(\psi_c) =
\begin{cases}
1 & \text{if } \bar{R}_{\psi_c} \geq \tau_q^{(l)}, \\
0 & \text{otherwise},
\end{cases}
\label{formula_m}
\end{equation}
where $\tau_q^{(l)}$ denotes the layer-wise pruning ratio (i.e., the fraction of components to be removed from each layer $l$), defined as the $q$-quantile of the relevance scores of components in that layer as follows:
\begin{equation}
\tau_q^{(l)} = \text{Quantile}_q\left( \left\{ \bar{R}_{\psi}^{(l)} \mid \psi \in \Psi^{(l)} \right\} \right).
\end{equation}
The binary mask $\mathcal{M}(\psi_c)\in\{0,1\}$ indicates whether component $\psi_c$ is retained ($\mathcal{M}(\psi_c)=1$) or removed ($\mathcal{M}(\psi_c)=0$). This ensures that the least relevant fraction $q$ of components in each layer is pruned. The choice of $q$ directly affects the trade-off between communication efficiency and generalization capability. A higher $q$ results in stronger pruning, reducing communication overhead while potentially degrading performance. Conversely, a lower $q$ preserves more parameters and maintains higher model performance, at the cost of reduced communication efficiency. After computing the pruning mask, we apply it to the global model parameters:
 \begin{equation}
w = w \odot \mathcal{M}(\psi_c),
\end{equation}
where $\odot$ denotes element-wise multiplication, and $\mathcal{M}(\psi_c)$ is aligned to match the dimensions of the weights associated with each pruned component. 

This yields a structured, sparsified global model that is then distributed to the clients for the next round of federated training. A crucial aspect of our strategy is the consistency and stability of pruning across communication rounds. To achieve this, once a pruning mask $\mathcal{M}$ is established, it is retained and reapplied in all following rounds, ensuring that all clients operate on structurally identical models. This design avoids structural divergence across clients, which could otherwise compromise the convergence of the federated optimization process. To avoid pruning before the global model stabilizes, the process is initiated only after completing a warm-up phase of $\upsilon$ communication rounds. This phase allows the global model to reach a relatively stable state where relevance scores become reliable indicators of component importance. After the warm-up, pruning is initiated and the relevance scores are periodically recomputed at the server side. This enables the pruning mask to adapt over time, reflecting changes in the global model’s behavior and maintaining the balance between model compression and task performance. By embedding explanation-guided pruning into the aggregation and distribution process, the proposed strategy substantially reduces communication overhead while ensuring that only task-relevant model components are retained and transmitted during the training. Our strategy is summarized in Algorithm \ref{alg:fl_lrp_pruning}.

\begin{algorithm}[ht]
\caption{The proposed FedX strategy}
\label{alg:fl_lrp_pruning}
\KwIn{
    $K$, $D_i$, $E$, $\alpha_i$, $\upsilon$, $q$, $M$, $R$, $D_\text{ref}$, $\phi$
}
\KwOut{$w$}
\SetKwFunction{FLocal}{LocalTraining}
\SetKwProg{PLocal}{function}{:}{end}

\SetKwProg{PAgg}{function}{:}{end}
\SetKwFunction{FLocal}{LocalTraining}
\SetKwProg{PLocal}{function}{:}{end}

\PLocal{\FLocal{$\mathcal{M}$, $w$}}{
$w_i \leftarrow w$\;
\For{$e \leftarrow  1$ to $E$}{
  \For{$\mathcal{B} \in D_i$}{
        Update $w_i$ by minimizing $\mathcal{L}$ on $\mathcal{B}$\;
    }
    
}$w_i \leftarrow w_i \odot \mathcal{M}$\;
}
 
 $\mathcal{M} \leftarrow \mathbf{1}$\;
$w \leftarrow w_0$\;
\For{ $r \leftarrow  1$ to $R$}{
    \If{$r = \upsilon$}{
        \For{each unit $\psi_c$ in $\phi$}{
             $\bar{R}_{\psi_c} \leftarrow \frac{1}{M} \sum_{z=1}^{M} R_{\psi_c}(I_z), $  \\
             where $I_i\in D_\text{ref}$\;
         }
         $ \mathcal{R}_q \leftarrow\{\bar{R}_{(\psi_1)}, \bar{R}_{(\psi_2)}, \ldots, \bar{R}_{(\psi_q)}\}$\;
         Compute $\mathcal{M}$ from $\mathcal{R}_q$ (see Eq. ~\ref{formula_m})\;
        
         $w^* \leftarrow w^* \odot \mathcal{M}$\;
        
    }
        \For{ $i \leftarrow  1$ to $K$}{
          $ w_i\leftarrow \FLocal(\mathcal{M},w^* ) $\;

}
    $w \leftarrow \sum_{i=1}^{K} \alpha_i w_i$\;
    }
\textbf{return}: $w$
\end{algorithm}

\section{Dataset Description and Design of Experiments}

\subsection{Dataset Description}

We conducted experiments on the BigEarthNet-S2 v2.0 \cite{clasen2024reben} and the EuroSAT \cite{helber2019eurosat} datasets. The BigEarthNet-S2 v2.0 dataset consists of 549,488 Sentinel-2 images acquired over 10 European countries. Each image is annotated with multiple labels based on the 19-class scheme from the CORINE Land Cover Map, as defined in \cite{sumbul2021bigearthnet}. The EuroSAT dataset consists of 27,000 Sentinel-2 images annotated with 10 classes. For BigEarthNet-S2, we utilized the summer subset, which comprises images from Austria, Belgium, Finland, Ireland, Lithuania, Portugal, Serbia and Switzerland. We followed the train–test split provided by \cite{clasen2024reben}. To simulate a decentralized, non-identically and independently distributed (non-IID) setting, we partitioned the training data such that each client was assigned data exclusively from a single country. As a result, the setup consisted of 8 clients, each representing a country and participating in every communication round. For EuroSAT, we used the RGB bands of the dataset. We followed the standard train–test split provided in the dataset. To simulate a decentralized, identically and independently distributed (IID) setting, we partitioned the training data randomly across clients, ensuring that each client received a representative subset of the overall data distribution. This resulted in a homogeneous setup, where all clients had access to similar class distributions. Employing a non-IID setting for BigEarthNet-S2 and an IID setting for EuroSAT enables the evaluation of the proposed strategy under varying levels of training data heterogeneity.

\subsection{Design of Experiments}
\label{doe}

In the experiments, as a FL algorithm, we exploited FedAvg \cite{li2019convergence}, which updates the global model by aggregating the weighted averages of locally trained models on clients. This algorithm was selected as the aggregation method, as it serves as a widely adopted baseline, allowing fair and interpretable comparisons between different pruning strategies. This choice allows us to ensure that performance differences are caused by the pruning method instead of the aggregation process. In the experiments, the number of communication rounds was set to 20, with each client performing 3 local training epochs per round to allow sufficient model updates while maintaining a manageable overall training duration. During the experiments, we observed that increasing the number of communication rounds or local epochs beyond this point did not lead to notable changes in model performance. We tested our strategy in the context of two learning tasks: 1) multi-label scene classification using BigEarthNet-S2; and 2) single-label scene classification using EuroSAT. We considered six different DL model architectures: 
ResNet6~\cite{he2016resnet} ($\approx 1.7$M parameters), 
ResNet18~\cite{he2016resnet} ($\approx 11.7$M parameters), 
ResNet50~\cite{he2016resnet} ($\approx 25.5$M parameters), 
ViT-B16~\cite{dosovitskiy2021an} ($\approx 86$M parameters), 
ConvNeXt V2 Atto~\cite{liu2022convnet} ($\approx 3.7$M parameters), and 
ConvNeXt V2 Nano~\cite{liu2022convnet} ($\approx 14.0$M parameters). As explanation-guided pruning methods, we selected LRP~\cite{yeom2021pruning} and IG~\cite{hatefi2024pruning} due to their proven success and reliability in RS scene classification \cite{klotz2025effectiveness}. We extended their applicability to FL by computing relevance scores to identify less important parameters and applying structured pruning to the global model before redistributing it to clients. For LRP hyperparameterization, each model was divided into four consecutive parts following the approach in \cite{hatefi2024pruning}. For the ResNet architectures, we used the epsilon rule in the low-level hidden layers, the gamma rule in the mid-level and high-level hidden layers, and the epsilon rule again in the fully connected layers. For ViT-B16, we employed the epsilon rule uniformly in all four parts. The relevance scores are computed only at the central server using a publicly available reference dataset $D_{ref}$, which is independent of the local data on clients. In our experiments, $D_{ref}$ is constructed by randomly sampling 500 images from the validation set. This reference set is used exclusively for server-side relevance estimation and pruning-mask generation, and it is not used for client-side training. Since the relevance computation is performed entirely at the central server, local data is not shared with the central server. We considered layer-wise pruning configuration, where model parameters are pruned independently within each layer, and a global pruning configuration, where the entire model is pruned at once. All models were trained using the Adam optimizer with a learning rate of 0.001, weight decay of 1e-4, and a batch size of 512. Pruning was applied after a warmup phase of 10 communication rounds (\(\upsilon=10\)). This choice reflects the trade-off between achieving model stability and minimizing communication overhead. Performance was evaluated using micro mean Average Precision (mAP) for the multi-label classification task and overall accuracy for the single-label classification task. We used mAP as it is well suited for multi-label classification tasks with class imbalance, providing a comprehensive measure by aggregating true positives, false positives, and false negatives across all classes.

\section{Experimental Results}

We carried out different kinds of experiments in order to: 1) analyze FedX on different pruning configurations and pruning methods; 2) evaluate FedX across different DL architectures; 3) assess the communication and computational efficiency of FedX across different DL architectures; and 4) compare FedX with the state-of-the-art FL algorithms.

\subsection{Analysis of FedX with Different Pruning Configurations and Pruning Methods }

\begin{figure}[h]
  \centering
  \includegraphics[width=\linewidth]{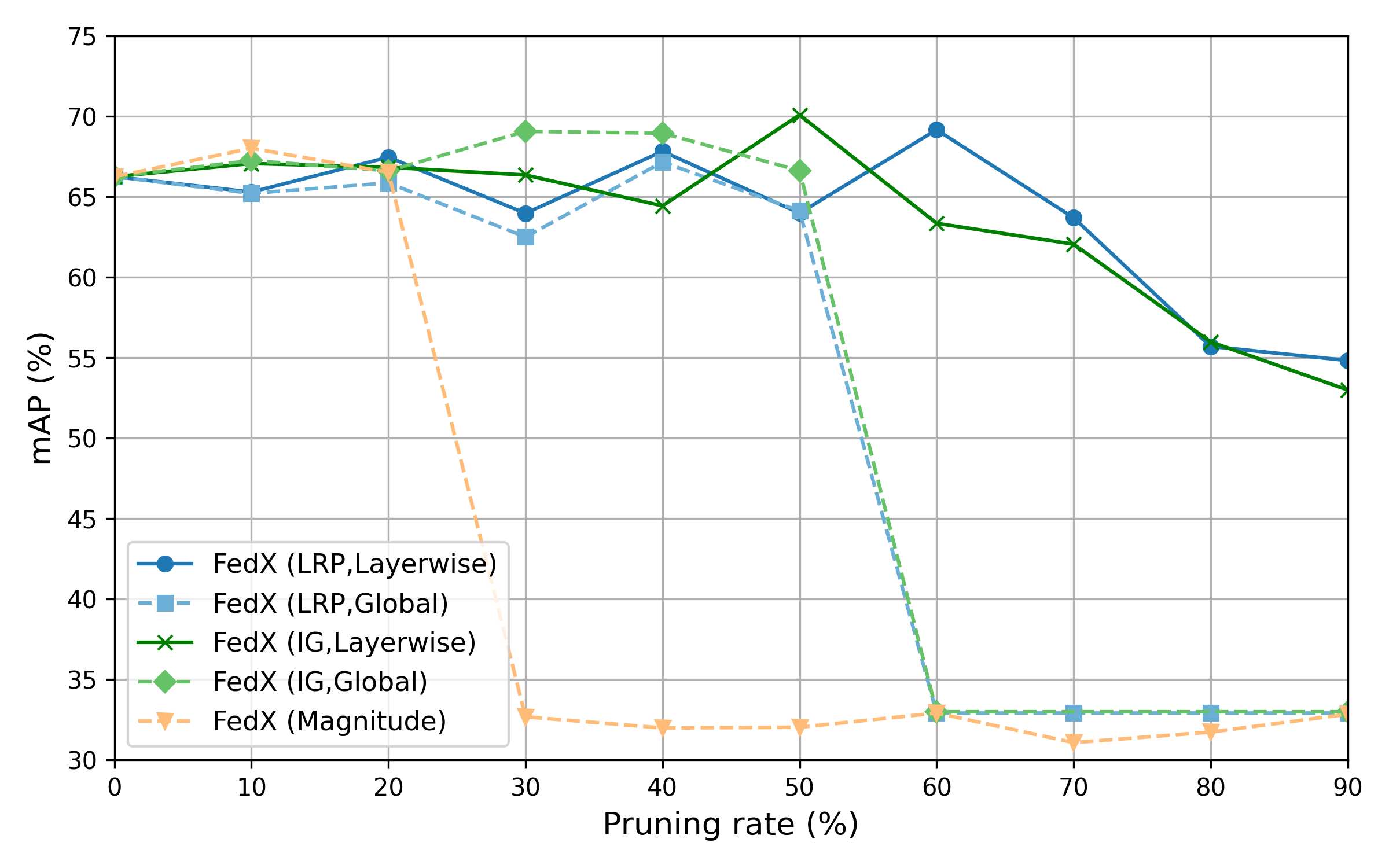}
  \caption{mAP scores obtained by FedX with different pruning configurations (layer-wise and global), relevance-based explanation methods (LRP and IG), and a magnitude-based pruning method for ResNet18 under varying pruning rates (the BigEarthNet-S2 dataset).}
  \label{fig:LRP_global_vs_layerwise_comparison}
\end{figure}

In this subsection, we assess the performance of FedX, considering different pruning configurations and pruning methods. To this end, we exploit ResNet18 as the backbone model due to its moderate number of parameters. As mentioned in Section \ref{doe}, we exploit LRP \cite{bach2015pixel} and IG \cite{sundararajan2017axiomatic} as explanation methods for the experiments, considering layer-wise and global pruning configurations. This results in five different scenarios: i) FedX with LRP and the layer-wise configuration, denoted as FedX (LRP, Layerwise); ii) FedX with LRP and the global configuration, denoted as FedX (LRP, Global); iii) FedX with IG and the layer-wise configuration, denoted as FedX (IG, Layerwise); iv) FedX with IG and the global configuration, denoted as FedX (IG, Global); and v) FedX with magnitude-based pruning, denoted as FedX (Magnitude), where components are pruned according to the L1 norm of the weights without relevance estimation. Fig.~\ref{fig:LRP_global_vs_layerwise_comparison} shows the obtained mAP scores for the BigEarthNet-S2 dataset.

\begin{figure}[h]
    \hspace*{-0.5cm}
    \includegraphics[width=1\linewidth]{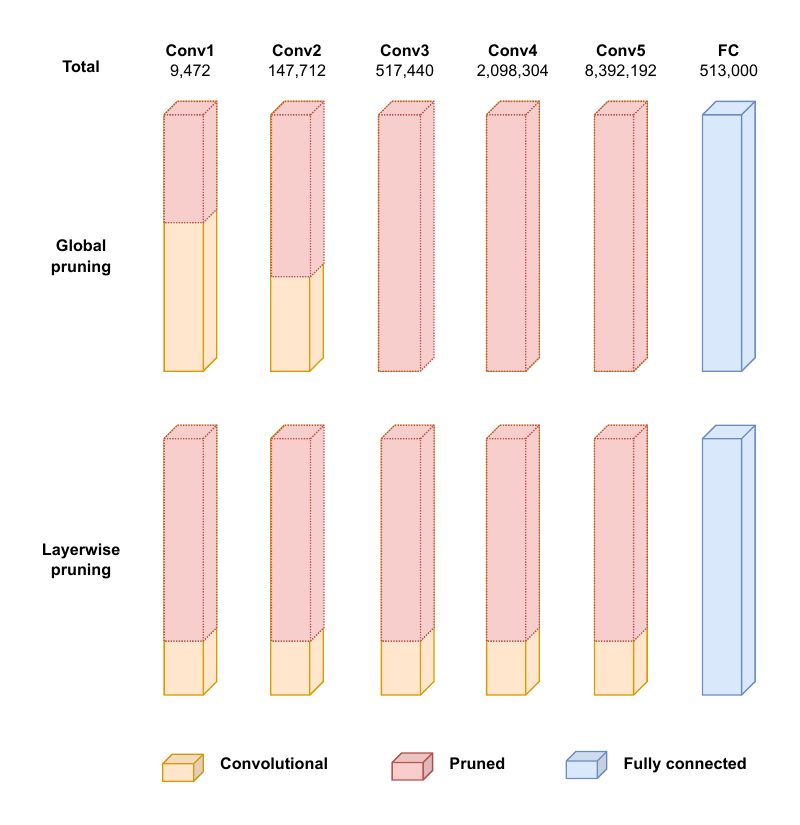}
    \caption{Distribution of retained and pruned parameters per layer in ResNet18 under 90\% overall sparsity, comparing global and layer-wise pruning configurations for both datasets. Yellow segments indicate retained convolutional parameters, red segments indicate pruned parameters, and the blue bar represents the fully connected layer (which is not subject to pruning).}
    \label{fig:lrp_vs_layerwise}
\end{figure}

As one can observe from the figure that both pruning configurations provide similar mAP scores when the pruning rate is between 10\% and 50\%. Within this range, the performance remains relatively stable with mAP values fluctuating between 64\% and 70\%. For example, at 20\% pruning, all scenarios achieve similar performance with mAP values around 67\%. However, once the pruning rate reaches 60\%, the performance of FedX (LRP, Global) and FedX (IG, Global) drops drastically to about 33\% and remains at this low level for higher pruning rates. In contrast, the layer-wise scenarios maintain much stronger performance with FedX (LRP, Layerwise) achieving 69\% and FedX (IG, Layerwise) reaching 64\% at the same pruning level. Even at 90\% pruning, FedX (LRP, Layerwise) and FedX (IG, Layerwise) retain 55\% and 53\%, respectively, while the global strategies remain at 33\%. This demonstrates that the layer-wise configuration is more robust, preserving model performance across a wide range of pruning rates, while the global configuration fails to sustain accuracy beyond moderate sparsity. This robustness can be explained by the structural behavior of LRP and IG. A key property of these methods is that the total relevance is conserved across layers, which means that each layer receives the same overall relevance regardless of its size \cite{bach2015pixel,sundararajan2017axiomatic}. In CNN architectures, deeper layers typically contain significantly more parameters than the earlier layers. However, due to relevance conservation, the average relevance per component becomes much lower in these deeper layers. When global configuration is applied, this imbalance introduces a bias against deeper layers, as components from all layers are compared directly. As a result, deeper layers are more likely to fall below the global pruning threshold and be removed, even when they contribute significantly to model performance. This effect is clearly illustrated in Fig.~\ref{fig:lrp_vs_layerwise}, where global pruning disproportionately removes parameters from the final convolutional blocks, while the earlier layers remain largely preserved. These results show that the layer-wise pruning configuration preserves model performance more effectively under extreme pruning, while also preventing the structural degradation observed with the global pruning configuration. Therefore, the layer-wise configuration offers a more reliable solution and supports higher pruning rates without a significant loss in accuracy. 


\begin{figure*}[t]
  \centering
  \begin{subfigure}[b]{0.49\linewidth}
    \includegraphics[width=\linewidth]{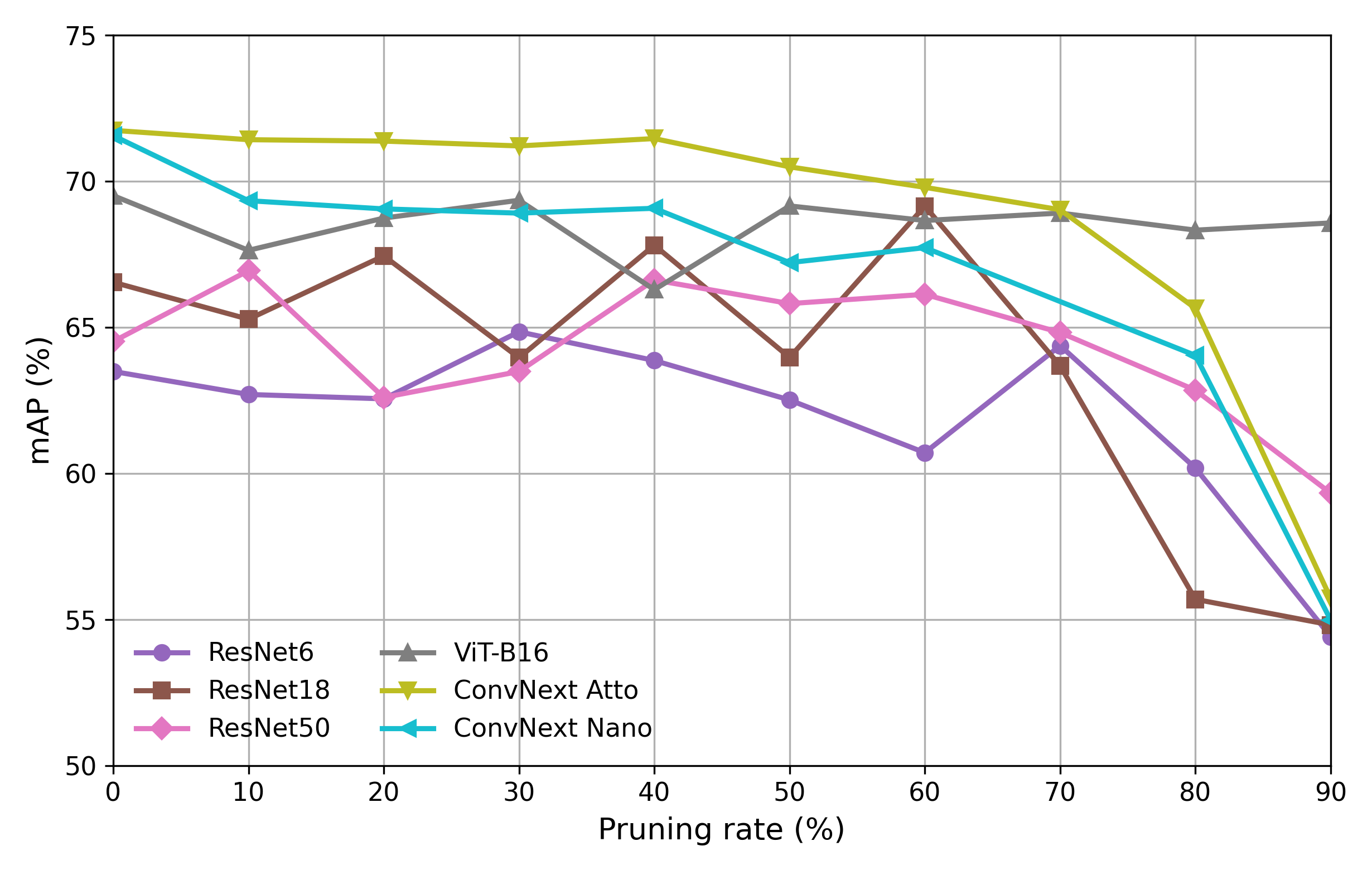}
    \caption{}%
    \label{fig:all_models_benv2}%
  \end{subfigure}%
  \hfill
  \begin{subfigure}[b]{0.49\linewidth}
    \includegraphics[width=\linewidth]{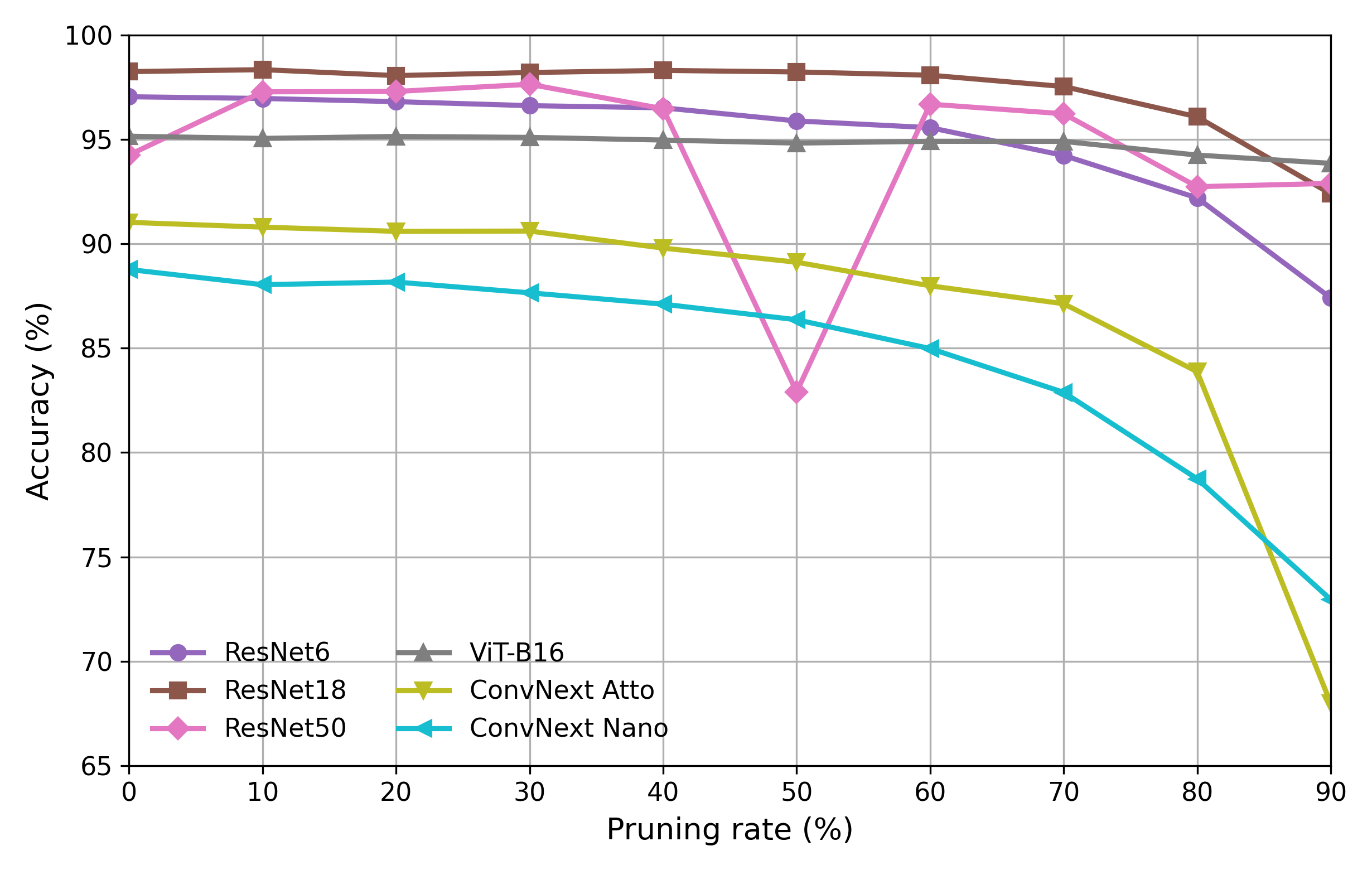}
    \caption{}%
    \label{fig:all_models_eurosat}%
  \end{subfigure}%
  \caption{Result scores obtained by FedX (using the LRP method with a layer-wise configuration) for different model architectures under varying pruning rates for: (a) the BigEarthNet-S2 (mAP); and (b) the EuroSAT (accuracy) datasets.}
  \label{fig:all_models_comparison}
\end{figure*}

From Fig.~\ref{fig:LRP_global_vs_layerwise_comparison}, one can also observe that FedX (IG, Layerwise) and FedX (LRP, Layerwise) provide similar mAP scores when the pruning rate is between 10\% and 40\%. Within this range, the performance remains relatively stable, with mAP values fluctuating between 63\% and 67\%. FedX (IG, Layerwise) performs slightly better than FedX (LRP, Layerwise) at pruning rates of 30\% and 50\%, reaching 66\% and 70\% respectively. However, FedX (LRP, Layerwise) outperforms FedX (IG, Layerwise) at pruning rates of 10\% and 60\%. As an example, when the pruning rate is set to 60\%, FedX (LRP, Layerwise) achieves an mAP score of 69\%, while FedX (IG, Layerwise) reaches 63\%. Once the pruning rate exceeds 70\%, the performance of FedX (IG, Layerwise) begins to degrade more rapidly than that of FedX (LRP, Layerwise). From this point onward, FedX (LRP, Layerwise) shows better robustness across the higher pruning range. At the highest pruning rate of 90\%, FedX (LRP, Layerwise) still retains 55\% mAP, whereas FedX (IG, Layerwise) drops further to 53\%. One can observe that despite the decline, FedX (LRP, Layerwise) retains a higher proportion of the original performance under extreme sparsity conditions. This behavior highlights the comparative robustness of FedX (LRP, Layerwise) in preserving relevant model structure when the number of model parameters is significantly reduced. While FedX (IG, Layerwise) shows better performance at certain pruning rates, FedX (LRP, Layerwise) performs better overall. In addition, FedX (Magnitude) performs comparably to the explanation-guided variants at low sparsity levels (up to 20\% pruning). However, its performance drops sharply beyond 30\% pruning and remains around 32\% mAP for higher pruning rates. This highlights that selecting components solely based on weight magnitude is not sufficient to preserve task-relevant capacity under stronger pruning, and confirms the benefit of explanation-guided relevance estimation in FedX. Based on its superior robustness and consistent performance, we use FedX (LRP, Layerwise) for the rest of the experiments, denoting it as FedX for simplicity.

\subsection{Performance Evaluation of FedX on Different DL Architectures}

In this subsection, we evaluate the effectiveness of FedX by considering different model architectures (ResNet6, ResNet18, ResNet50, ConvNeXt Atto, ConvNeXt Nano, and ViT-B16). Fig.~\ref{fig:all_models_comparison} presents the mAP scores on BigEarthNet-S2 and the accuracies on EuroSAT obtained by FedX with pruning rates ranging from 0\% to 90\%.

From Fig.~\ref{fig:all_models_comparison}a, one can observe that all models exhibit a degree of performance degradation with increasing sparsity. The relatively flat performance curve of ViT-B16 highlights the strength of FedX in preserving task-relevant information even under extreme compression. As an example, FedX maintains mAP values above 68\% throughout most of the pruning range using ViT-B16. This is mainly due to its larger parameter capacity, which enables ViT-B16 to preserve essential representations even after strong pruning. Similarly, FedX retains mAP above 65\% up to 80\% pruning, with its performance only dropping more significantly at the final sparsity level using ConvNeXt Atto. These trends confirm that FedX is highly compatible with modern high-capacity architectures that benefit from structured attention or depthwise convolutional modules. Moreover, FedX enables competitive performance preservation among CNNs. As an example, ResNet50 maintains a relatively stable curve compared to ResNet6 and ResNet18, indicating that deeper models benefit more from relevance-guided pruning. Although FedX exhibits the steepest decline, it still reaches competitive performance at 90\% pruning rate using ResNet6. This shows that even low-capacity models can be pruned aggressively without complete failure. These results suggest that FedX not only scales well across different architectural designs, but also effectively prevents abrupt accuracy collapse at high sparsity levels. This robustness is particularly valuable in FL settings, where communication and memory constraints require strong model compression. The ability to retain high accuracy across pruning rates confirms that FedX generalizes well across a broad spectrum of model families. Another notable observation from the figure is that FedX enables pruned models to achieve performance that can even surpass their original unpruned versions. As an example, ConvNeXt Atto maintains nearly the same mAP score between 30\% and 70\% pruning, with small fluctuations where the pruned versions slightly outperform the model with no pruning. A similar trend is observed in ViT-B16, where the model reaches its peak mAP at 30\% pruning, surpassing the mAP score of its unpruned version. This suggests that FedX not only preserves essential representations, but may also eliminate redundant parameters, leading to improved generalization. These findings indicate that relevance-guided pruning can act as an effective structural regularizer, particularly in over-parameterized architectures, enabling them to operate more efficiently without compromising model performance.

From Fig.~\ref{fig:all_models_comparison}b, one can observe that FedX consistently maintains strong performance across various architectures as sparsity increases. In particular, it demonstrates robustness, maintaining an accuracy above 95\% across nearly the entire pruning range, with minimal degradation even at high sparsity levels. This suggests that FedX effectively leverages the representational capacity of mid-sized CNNs to preserve essential features. Moreover, FedX maintains around 94\% accuracy at all pruning rates with ViT-B16, showing strong compatibility with transformer-based models and resilience to accuracy drops. The performance decline is more gradual with ConvNeXt Atto and ConvNeXt Nano, especially beyond 60\% pruning. However, both models still retain competitive accuracy of up to 80\%. FedX leads to a temporary drop in performance at 50\% pruning using ResNet50. This suggests that while some redundancy may initially be removed, the model can reallocate relevant features as pruning progresses. Even with the smallest architecture (i.e., ResNet6), FedX maintains reasonable accuracies up to 70\% sparsity, showing its effectiveness in highly constrained settings. These results indicate that FedX scales well with both compact and large models, enabling aggressive pruning without critical loss in performance. Furthermore, several models, such as ResNet50 and ConvNeXt Atto, reach their highest accuracies at intermediate pruning levels. This suggests that FedX may improve the generalization capability of the global model.

\subsection{Communication and Computational Efficiency of FedX on Different DL Architectures}

In this subsection, we analyze the communication and computational efficiency of FedX across different model architectures. Table~\ref{tab:communication_cost_saved} shows the total communication cost of the unpruned model and the communication cost reductions achieved by pruning. The total cost values represent the cumulative uplink and downlink communication of the model parameters between the clients and the server. As shown in the table, higher pruning rates lead to greater communication cost reductions for all architectures due to the reduced transmission of model parameters. One can observe that models with larger parameter sizes benefit more significantly from pruning in absolute terms. As an example, ViT-B16 yields a total communication cost of approximately 106.4 GB without pruning. At 90\% pruning, the communication cost reductions reach 46.3 GB, which corresponds to a reduction of nearly 44\% of the total communication volume. Similarly, ConvNeXt Nano and ResNet50 achieve communication cost reductions of 8.2 GB and 12.9 GB, respectively, under the same pruning level. Even compact models such as ResNet6 and ConvNeXt Atto benefit from FedX. At 90\% pruning, ResNet6 saves 183 MB, nearly 44\% of its original 412 MB, while ConvNeXt Atto reduces its cost by 1.85 GB from an initial 4.1 GB. These results indicate that FedX significantly reduces communication costs across a diverse set of architectures, regardless of their size or depth. The ability to maintain accuracy, as shown in earlier sections, while cutting communication by up to 40–45\% highlights the benefit of this strategy in bandwidth-constrained FL scenarios.

In addition to communication efficiency, we also assess the computational cost caused by the relevance estimation and pruning processes. Table~\ref{tab:compute_runtime} summarizes the floating-point operation counts (FLOPs) for both the relevance estimation and pruning mask application steps. These measurements reflect the cost of a single pruning cycle per client and are obtained on GPU hardware tailored to each architecture. The relevance estimation step is the more computationally intensive phase due to the need to propagate relevance backward through the network. Among the models evaluated, ViT-B16 exhibits the highest computational demand, which requires $5.0 \times 10^{13}$ FLOPs for the relevance computation. This is followed by ResNet50, which requires $4.38 \times 10^{12}$ FLOPs for relevance estimation. In contrast, lighter models such as ResNet6 and ConvNeXt Atto require substantially fewer FLOPs, reflecting their lower complexity. In particular, the pruning mask generation step is extremely lightweight across all models, requiring more than $10^{5}$ times fewer FLOPs than the relevance estimation step. These findings demonstrate that FedX is both communication-efficient and computationally practical. Although the relevance estimation phase introduces moderate computational cost, it occurs only once per pruning round and is vastly outweighed by the communication cost reductions over the course of training. This balance makes the strategy well-suited for deployments where uplink and downlink costs are critical, and computational capacity at the edge remains limited but sufficient.

\begin{table*}[h]
\centering
\caption{Communication cost of unpruned training and amount of data saved through pruning at different sparsity levels obtained by FedX (using the LRP method with a layer-wise configuration) across different architectures on both datasets.}
\label{tab:communication_cost_saved}
\begin{tabular}{llr*9{r}}
\toprule
\multicolumn{2}{c}{\textbf{Model}} & \textbf{Total Cost} & \multicolumn{9}{c}{\textbf{Saved Cost (MB) per Pruning Rate}}\\
\cmidrule(lr){1-2}\cmidrule(lr){3-3}\cmidrule(lr){4-12}
Name & Size (MB) & Unpruned (MB) & 10 & 20 & 30 & 40 & 50 & 60 & 70 & 80 & 90\\
\midrule
ResNet6  & 1.29  & 412  & 19 & 39 & 60 & 80 & 102 & 121 & 141 & 162 & 183 \\
ResNet18  & 42.76  & 13682  & 675 & 1358 & 2038 & 2721 & 3416 & 4091 & 4774 & 5454 & 6137 \\
ResNet50  & 89.91  & 28771  & 1419 & 2856 & 4283 & 5720 & 7172 & 8592 & 10028 & 11456 & 12893 \\
ConvNeXt Atto  & 12.96  & 4148  & 205 & 411 & 617 & 823 & 1029 & 1234 & 1440 & 1646 & 1852 \\
ConvNeXt Nano  & 57.24  & 18315  & 912 & 1824 & 2736 & 3648 & 4561 & 5473 & 6385 & 7297 & 8210 \\
ViT-B16  & 332.6  & 106432  & 5147 & 10294 & 15441 & 20588 & 25735 & 30882 & 36029 & 41176 & 46323 \\

\bottomrule
\bottomrule
\end{tabular}
\end{table*}

\begin{table}[h]
  \centering
  \caption{Average computational cost in FLOPs for relevance estimation and pruning mask application obtained by FedX (using the LRP method with a layer-wise configuration) across different architectures on both considered datasets.}
  \label{tab:compute_runtime}
  \begin{tabular}{l|cc}
    \toprule
    \textbf{Model} & \textbf{Relevance Estimation} & \textbf{Pruning Mask Application} \\
    \midrule
    ResNet6 & $1.20\times10^{12}$ & $1.01\times10^{6}$ \\
    ResNet18 & $2.49\times10^{12}$ & $3.36\times10^{7}$  \\
    ResNet50 & $4.38\times10^{12}$ & $7.05\times10^{7}$  \\
    ConvNeXt Atto & $5.67\times10^{11}$ & $1.01\times10^{7}$  \\
    ConvNeXt Nano & $2.49\times10^{12}$ & $4.48\times10^{7}$ \\
    ViT-B16 & $5.00\times10^{13}$ & $1.07\times10^{8}$ \\

    \bottomrule
  \end{tabular}
\end{table}

\subsection{Comparison of FedX with State-of-the-Art Algorithms}

\begin{figure}[t]
  \centering
  \includegraphics[width=\linewidth]{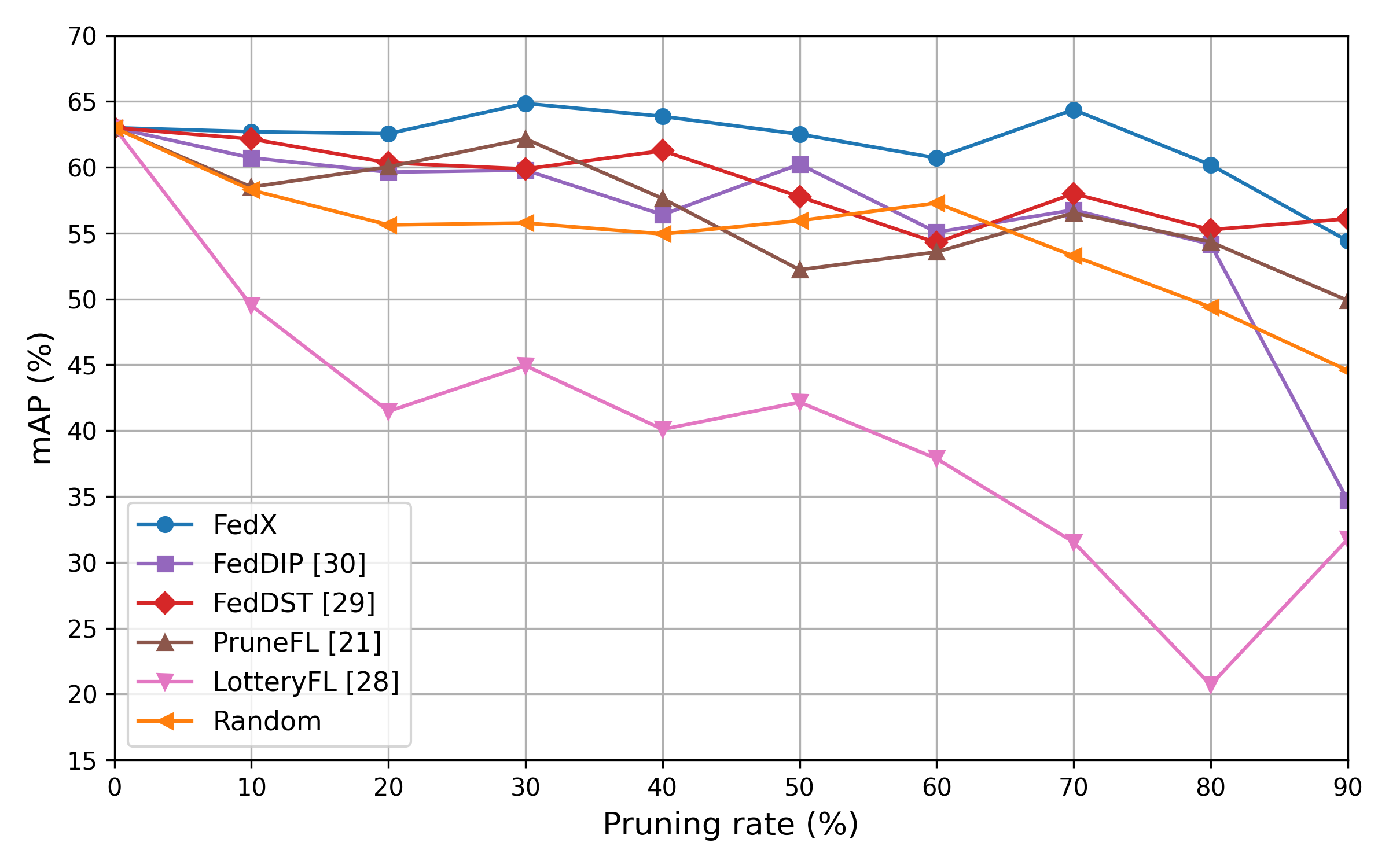}
  \caption{mAP scores obtained by FedX (using the LRP method with a layer-wise configuration), FedDST, FedDIP, PruneFL, LotteryFL, and random pruning for ResNet6 under varying pruning rates (the BigEarthNet-S2 dataset).}
  \label{fig:resnet6_lrp_comparison}
\end{figure}

In this subsection, we compare FedX with random pruning and four state-of-the-art algorithms: FedDST \cite{bibikar2022federated}, FedDIP \cite{long2023feddip}, LotteryFL \cite{li2021lotteryfl}, and PruneFL \cite{jiang2022model}. The comparison is carried out on different pruning rates using the ResNet6 architecture. This architecture was chosen due to its relatively small number of parameters, which limits its capacity to compensate for pruning through overparameterization. This choice also ensures a fairer comparison of pruning strategies and allows a clearer understanding of their impact on recovery and convergence under high sparsity. Fig.~\ref{fig:resnet6_lrp_comparison} shows the corresponding mAP scores for the BigEarthNet-S2 dataset.

From the results, one can see that FedX consistently yields higher mAP values than random pruning and the considered state-of-the-art algorithms across almost all sparsity levels. As an example, at a pruning rate of 20\%, FedX achieves an mAP score of 64\%, outperforming the randomly pruned model by 6\% and exceeding FedDST, FedDIP, and PruneFL by approximately 4\%. In greater detail, as the pruning rate increases from 20\% to 50\%, all the considered methods show a gradual decline in mAP. However, FedX achieves the highest performance throughout this range. At the 50\% pruning rate, FedX achieves an mAP score of 63\%, outperforming random pruning by 7\%, FedDIP by 3\%, FedDST by 5\%, PruneFL by 10\%, and LotteryFL by 20\%. Beyond this point, the performance gap widens with FedX, especially at extreme sparsity. As an example, at a pruning rate of 80\%, FedX reaches 60\% mAP, outperforming random pruning by 11\% and FedDST, FedDIP, and PruneFL by around 5\%, while the performance gap compared to LotteryFL is substantially larger at this pruning level. Even at a 90\% pruning rate, FedX maintains competitive performance, while the comparison methods degrade substantially. This indicates that FedX is more effective than the state-of-the-art algorithms and random pruning in preserving essential model components, particularly in scenarios where model capacity is severely constrained. Across all pruning rates, FedX consistently maintains performance closer to the unpruned model compared to the random pruning and state-of-the-art algorithms. This shows that FedX can maintain performance remarkably close to the unpruned model, even at extreme sparsity rates. These results demonstrate the robustness of FedX in FL settings with constrained communication and model size, and highlight its ability to guide pruning in a performance-preserving manner. A similar behavior was also observed by using the EuroSAT dataset (not reported due to space constraints).

\section{Conclusion}

In this paper, we have presented FedX, an explanation-based pruning strategy to reduce the communication overhead in FL for RS image classification. FedX utilizes backpropagation-based explanation methods to guide pruning decisions based on the importance of model parameters. By applying relevance estimation at the server side and pruning less informative components from the global model, our strategy significantly reduces the volume of global model parameters exchanged between clients and the central server during training. This ensures that model updates retain task-relevant information while achieving substantial communication efficiency.

For the experiments, we first evaluated the impact of different explanation methods (LRP and IG) and pruning configurations (global and layer-wise). The results show that LRP consistently outperforms IG, particularly under high sparsity levels. Moreover, layer-wise configuration achieves higher model accuracy than global configuration by avoiding the structural imbalance introduced by uniform relevance thresholds across layers. This confirms that preserving the relative importance of parameters within each layer is critical to maintain performance when the pruning rate is very high. These findings highlight the importance of method selection and configuration when applying explanation-based strategies in FL. We also studied the influence of pruning strategies across communication rounds. FedX demonstrated more stable convergence and stronger recovery compared to state-of-the-art methods and random pruning at different pruning levels. Even under high sparsity, the pruned models achieved performance levels close to, or exceeding, the unpruned baseline after a limited number of communication rounds. These results show that FedX not only compresses models effectively, but also facilitates fast and reliable convergence in FL training. The effectiveness of FedX was further validated across a diverse set of DL architectures. Our evaluation reveals that modern architectures are more resilient to high pruning rates, and in several cases, the pruned models even surpass their unpruned versions in classification performance. This behavior is particularly evident in ViT models, where large parameter capacity allows the network to retain key representations even after pruning. The consistent performance across different architectures confirms the general applicability and robustness of FedX.

In addition to performance improvements, we analyzed the communication and computational costs of FedX. Quantitative results indicate that communication cost reductions scale proportionally with the pruning rate and are particularly significant for large models. For instance, FedX reduces total communication by up to 44\% while maintaining high accuracy. Although relevance estimation introduces moderate computational overhead, this process is performed once per model training and remains lightweight compared to the overall training cycle. The pruning step itself is computationally efficient and does not introduce additional complexity to client-side operations. We would like to emphasize that FedX is architecture-agnostic and does not require any modifications to the client-side training pipeline. It operates entirely on the server side and can be seamlessly integrated into existing FL workflows. This flexibility makes it suitable for large-scale RS applications, where data privacy, legal restrictions, and communication cost are major constraints.

FedX demonstrates strong performance in various settings. However, it applies the same sparsity pattern across all communication rounds and clients. In FL, where data distributions on clients are heterogeneous, tailoring pruning decisions dynamically could further enhance the generalization capability of the global model. As a future work, we plan to explore dynamic pruning schedules that adapt sparsity levels based on model performance. Additionally, we aim to investigate client-specific relevance estimation to account for data heterogeneity.

\bibliographystyle{IEEEtran}
\bibliography{main}
\end{document}